\pdfoutput=1

\documentclass[11pt]{article}

\usepackage{EMNLP2023}

\usepackage{times}
\usepackage{latexsym}
\usepackage{array}
\usepackage{float}
\usepackage{makecell}
\usepackage[T1]{fontenc}

\usepackage[utf8]{inputenc}

\usepackage{microtype}

\usepackage{inconsolata}
\usepackage{amsmath}
\usepackage{amssymb}
\usepackage{graphicx}

\title{Infor-Coef: Information Bottleneck-based Dynamic Token Downsampling for Compact and Efficient language model}

\author{{\bf Wenxi \enspace Tan}\\
	Fudan University\\
    \texttt{tanwww1229@gmail.com}}
\begin{document}
	\maketitle
	\begin{abstract}
		The prevalence of Transformer-based pre-trained language models (PLMs) has led to their wide adoption for various natural language processing tasks. However, their excessive overhead leads to large latency and computational costs. The statically compression methods allocate fixed computation to  different samples, resulting in redundant computation. The dynamic token pruning method selectively shortens the sequences but are unable to change the model size and hardly achieve the speedups as static pruning. In  this paper, we propose a model accelaration approaches for large language models that incorporates dynamic token downsampling and static pruning, optimized by the information bottleneck loss. Our model, Infor-Coef,  achieves an 18x FLOPs speedup with an accuracy degradation of less than 8\% compared to BERT. This work provides a promising approach to compress and accelerate transformer-based models for NLP tasks.
	\end{abstract}
	
	\section{Introduction}
	Large language models based on Transformer \citep{attention} architectures, such as BERT \citep{bert}, RoBERTa \citep{liu2020roberta} , and GPT models (\citealp{gpt2}, \citealp{gpt3}), have gained prominence in recent years for their remarkable state-of-the-art performance in various tasks related to Natural Language Processing (NLP). These works rely on deep networks with millions or even billions of parameters, and the availability of high computation and large storage capability plays a key role in their success. In this regard, there has been a proliferation of studies aimed at improving the efficiency of large language models, including knowledge distillation (\citealp{distill}, \citealp{distilbert}, \citealp{tinybert}), quantization \citep{qbert}, low-rank factorization\citep{matrix}, weight sharing \citep{albert}, and weight pruning (\citealp{movement}, \citealp{CoFi}) and dynamic accelerating (\citealp{deebert}, \citealp{power}). 

Pruning has emerged as a promising approach to compress and accelerate DNN models, significantly reducing storage and computational costs. Structured pruning method delivers a static compact model by removing structured blocks of weights, e.g. heads (\citealp{heada}, \citealp{headb}) and encoder layers \citep{Reducing}. However, removing a large proportion of parameters may result in noticeable accuracy loss. To address this, the distillation paradigm is commonly adopted for recovery training, where the pruned model learns the knowledge delivered from the unpruned model. (\citealp{movement}) While these pruning methods achieve compelling results, they are static and have a fixed computation route for all inputs, regardless of the differing information redundancy of various sequences. 

Another pruning approach that we consider in this paper is token pruning, a dynamic pruning method that reduces computation by progressively dropping unimportant tokens in the sequence, allocating adaptive computational budget to different samples. It is appealing for its similarity to humans, who pay more attention to the more informative tokens. 

We draw inspiration from the work of \citep{power} who demonstrated that attention-based models accumulate information redundancy as tokens pass through encoder layers.  Based on this observation, we propose a dynamic pruning method that downsamples tokens before each encoder layer, in accordance with  an information compression demand. To deploy and optimize this compression process, we utilize the information bottleneck (IB) principle \citep{IB}.  IB recognizes the deep neural network as a process of information compression and extracting, optimized by maximizing the mutual information of inputs and labels, while controlling the mutual information between the inputs and hidden representatives. \citep{blackbox} We explore the potential of applying IB principle on token pruning. However, thus far, token pruning method rarely achieves large speedups (1.5-3x at most) as it leaves the model parameters intact ( \citealp{LTP} ) or introduces additional parameters (\citealp{Transkimmer}, \citealp{trbert}). In this work, we propose Infor-Coef, combining the \textbf{infor}mation bottleneck-based token downsampling with static pruning to create a highly\textbf{ co}mpact and \textbf{ef}ficient model. 

Our empirical results on the GLUE benchmark demonstrate that Infor-Coef outperforms different static pruning, dynamic pruning, and distillation baselines at various levels of speedups, with slight accuracy degradation of less than 8\%. Specifically, Infor-Coef achieves 18x FLOPs speedup with padding and 16x reduction without extra padding tokens. We also show that our IB-based optimization yields better results than the typical $l0$-norm-based token pruning loss function.
\footnote{\url{https://github.com/twwwwx/Infor-Coef}}	
	
	\section{Related Works}
	\subsection{Structured Pruning with Distillation}
	Pruning searchs for a compact subnetwork from an overparameterized model by eliminating the redundant parameters and modules. Different pruning granularities, from fine-grained to coarse-grained, include unstructured pruning by removing individual weights (\citealp{LTH},\citealp{movement}, \citealp{movement}), head pruning in multihead attention mechanism (\citealp{heada},\citealp{headb}), intermediate dimension pruning in feed-forward layer (\citealp{mccarley2021structured},\citealp{dynabert}), and entire encoder unit dropping \citep{Reducing} have been investigated to reduce the model size. Among them, unstructured pruning yields irregular weights elimination and won't necessarily boost efficiency.
	Structured pruning, targeted at reducing and simplifying certain modules and pruning structured blocks of weights, delivers compact models and achieves speedup.
	
	Distillation is applied to transfer the knowledge from the larger model to a smaller model. \citep{distill} Distillation objective is commonly adopted and leads to significant performance improvements for training during or after pruning. (\citealp{block},\citealp{movement}) 
	
	The unified structured pruning framework, CoFi \citep{CoFi}, jointly prunes different granularity of units while distilling from predictions and layer outputs to maintain the performance. It prunes 60\% of the model size without any accuracy drop. 
	
	\subsection{Dynamic Token Pruning}
	
	Unlike the static pruning strategy with a fixed computation cost, dynamic compression strategies are devised to selectively and adaptively allocate computation conditioned on different inputs. The dynamic approaches include dynamic depth \citep{deebert}, dynamic width \citep{ebert} and dynamic token length. Dynamic token length method accelerates the Transformer model by progressively dropping the tokens of less importance during inference. PoWER-BERT \citep{power}, one of the earliest works, recognizes the tokens as redundant for pruning. This is extended by LAT\citep{LAT} which uses LengthDrop, a skimming technique to drop tokens and recover them in the final layer, followed by an evolutionary search. Learned Token Pruning \citep{LTP} improves PoWER-BERT by introducing soft thresholds optimized in training.  However, as is discussed in \citep{trbert}, their attention weights-based token pruning strategies can lead to a suboptimal selection. TR-BERT \citep{trbert} adopts reinforcement learning on token skimming but is hard to converge. Transkimmer  \citep{Transkimmer} exploits a parameterized module that functions as token selector before each encoder layer that can be optimized using reparameterization trick. 

	\subsection{Information Bottleneck Principle}
	Information bottleneck (IB) was first proposed in \citep{IB}. IB principle can be utilized to interpret and analyze the deep neural networks  \citep{IBlearning}. VIB \citep{VIB} extends it by presenting a variational approximation to get a tractable bound and leverage backpropagation in training. Originally, information bottleneck theory takes the internal representation of the intermediate layer as hidden variable $Z$ of the input variable $X$. It aims to extract the representation $Z$ of $X$ that pertains the mutual information $I(X;Y)$ between the original input and target output, as well as compresses the mutual information $I(X;Z)$.
\begin{figure*}
		\centering
		\includegraphics[scale=0.6]{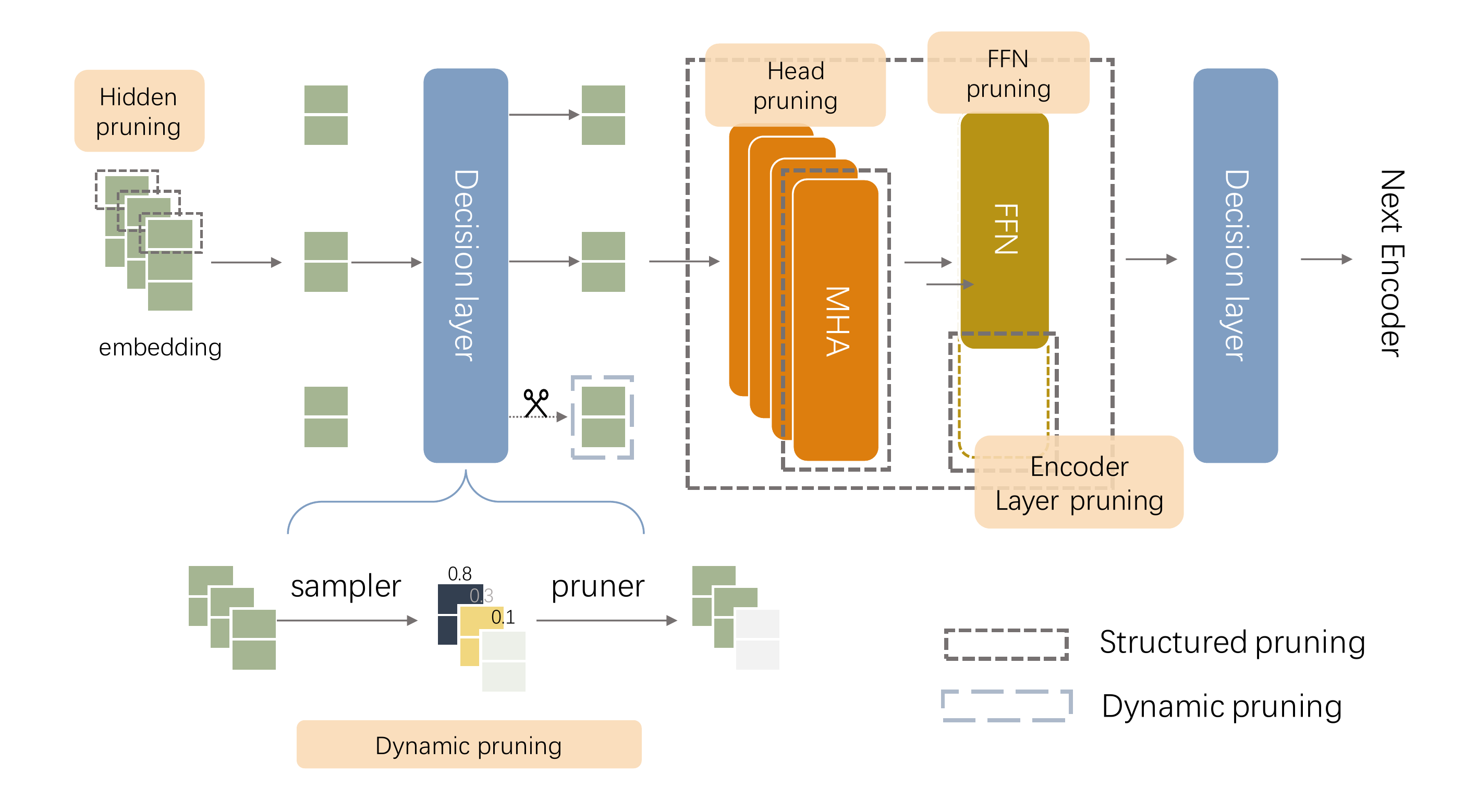}
		\caption{Overview of Infor-Coef. The dotted bordered rectangle denotes that the units / hidden dimensions are pruned using different kinds of masks. The structured pruning masks and dynamic pruning masks are learned using distillation objectives and information bottleneck respectively.}
		\label{overview}
\end{figure*}	
	In \citep{compressIB}, the successive intermediate representations are regarded as a Markov train, then IB is used to penalize model weights and delivers statically pruned  LeNet \citep{lenet} and VGG models \citep{VGG}.  To the best of our knowledge, the method proposed in this work is the first to explore IB principle in terms of dynamic token pruning.	
	\section{Methodology}
	We propose a collaborative pruning strategy, Infor-Coef, that implements static model pruning (section \ref{3.2}) and performs dynamic token downsampling (section \ref{3.3}) with a variational information bottleneck objective (section \ref{3.4}). We depict the overview of our model structure in Figure \ref{overview}.

	\subsection{Static Pruning\label{3.2}}
	
	The weights and computations of transformer \citep{attention} model mainly come from $H$ (e.g. 12) layers of multihead attention (MHA) and feed-forward network (FFN) modules. The embedded sequence matrix $x \in \mathbb{R}^{L}\times \mathbb{R}^d$, where $L$ corresponds to the token length and $d$ to the feature dimension (which is usually equal to 768 in BERT models). 
	
	Inside BERT, an MHA layer with $N_h$ (e.g.12) heads process the input sequence in parallel.
	After the MHA layer, the FFN layer follows, which first projects the processed sequence into a hidden size of $F$, and then down projects it to the original size to facilitate addition with the residual connection. In the static slenderization, we systematically reduce both the depth ($H$) as well as the width ($N_h$,$F$,$d$) of the model.
	
	We leverage the pruning and distillation strategy from CoFi \citep{CoFi}. Specifically, we exert masks with different positions and granularity of 
	(1) the feature dimension $d$; 
	(2) heads in the MHA layer;
	(3) intermediate dimension $F$ in FFN layer;
	(4) the entire MHA layer;
	(5) the entire FFN layer.
	
	Following \citep{l0} and \citep{structured}, we generate hard concrete distributions to leverage the $l0$ regularization. In the forward pass, masks are sampled to prune the corresponding neurons and get the overall sparsity $s$. Given a predefined sparsity ratio $\hat{s}$, the $l0$ penalty is 
	\begin{equation}
		\mathcal{L}_{0} = \mu_1 (\hat{s}-s) + \mu_2 (\hat{s}-t)^2  \label{static loss}
	\end{equation}
	where  $\mu_1$ and $\mu_2$ are lagrangian multipliers that are updated during training to push the model towards a fixed size.
	
	Since the removal of weights may lead to large performance degradation, distillation objectives are also added. We implement both layerwise distillation and output prediction distillation in \citep{CoFi} from the original model and the pruned model.

	\subsection{Dynamic Token Downsampling \label{3.3}}
	The hidden representation of a sentence in a MHA layer undergoes inner product operations along the dimension of the sentence's length in a self-attention mechanism, thus leading to a computational complexity that is almost proportional to the square of the sentence's length. With the inputs varying in complexity, we use dynamic token downsampling for sample-wise length reduction before each MHA layer.
	
	We adopt the MLP decision layer and reparameterization trick in \citealp{Transkimmer}.
	
	\subsubsection{Token Sampler}
	To achieve the hierarchical token elimination, we sample binary masks corresponding to each token in each encoder layer. 
	
	Let $h^i\in \mathbb{R}^{L^i}\times \mathbb{R}^d$ denote the $i,i\in 1,\dots,H$th hidden state. Before entering the $i$th encoder layer, it is passed through a sampling module $Sampler_i$, which generates the likelihood of "pruning" each token with probabilities $\pi^i \in [0,1]^{L^{i}}$ and samples $\textbf{}{z}^i \in \{0,1\}^{L^i}$ accordingly. 
	Following \citep{Transkimmer} and TR-BERT\citep{trbert}, the $Sampler$ is set to be a two-layer  MLP function. It always makes the "no pruning" decision at the initial state. We forward the outputs of it to the softmax function to get a Bernoulli parameter:
	\begin{equation}
		\begin{aligned}
			&(\pi_0,\pi_1)=softmax(MLP(x))\\
		\end{aligned}
	\end{equation}
	The probability of pruning the token $\pi_0$ is also used in the loss function, which we would explore in section \ref{3.4}.
	
	The discrete binary masks are not differentiable. For optimization, we take the reparameterization method, approximating the Bernoulli distribution with the Gumbel-Softmax trick. \citep{gumbel}
	Now the sampler is:
	\begin{equation}
		\begin{aligned}
			z&=Gumbelsoftmax((\pi_0,\pi_1)\\
			&=one\_hot( \mathop{\arg\max}\limits_{i\in\{0,1\}}[g_i+\log \pi_i])
		\end{aligned}
	\end{equation}
	where $g_i$ is drawn from $Gumbel(0,1)$. For differentiating, Gumbel-Softmax trick replaces the argmax operation with a softmax function.
	\subsubsection{Token Pruner}
	Now we get the pruned hidden states with $s^i=Pruner(h^i,\textbf{z}^i)$, at length $L^{i+1}$.
	During inference we actually prune certain tokens for $z_l^i = 0$ in the $Pruner(h^i,\textbf{z}^i)$, so $L^i\ge L^{i+1}$.But during training, we only set the pruned tokens zeroed out to simulate the pruning process, so theoratically we have $s^i=diag\{\textbf{z}^i\}h^i$, $L^i \equiv L$. 
	
	In the operation, we do not directly mask the tokens, considering that the zeroed token would affect other tokens in the self-attention mechanism.
	Instead, we convert the token masks to attention masks by:
	\begin{equation}
		\begin{aligned}
			R&=\exp(\dfrac{Q K^T}{\sqrt{d_h}})\\
			M_{ij} &= \mathbb{I}_{z_i=1}\mathbb{I}_{z_j=1}\\
			Attn' &= \dfrac{M_{ij}R_{ij}}{\sum_{i=1}^L M_{ij}R_{ij}} 
		\end{aligned}
	\end{equation}
	where $Q,K,V$ denote the query, key, value matrix respectively, and $d_h$ stands for the head size. $\mathbb{I}_{p}$ equals 1 given p is true, otherwise $\mathbb{I}_{p} = 0$.
	
	In this way, we eliminate the effects and cut off the information flow with regard to the masked tokens. Additionally, the pruned tokens in the downsampling are forwarded to the last hidden layer, which is the same as LAT \citep{LAT} and Transkimmer\citep{Transkimmer}.
	
	\subsection{Variational Information Bottleneck \label{3.4}}
	In this section, we introduce a variational information bottleneck loss to guide the information flow in token downsampling. Basically, we minimize the mutual information before and after the downsampling, while maintaining the mutual information between the preserved tokens and the true labels.
	\subsubsection{variational approximation}
	We use the same notations in section \ref{3.2} and section \ref{3.3}. Hence, $p(s^i|h^i)$ is defined via the relation
	\begin{equation}
		\begin{aligned}
			s^i&= Pruner(h^i,\bold{z}^i) \\
			&=diag \{\bold{z}^i\}h^i  \\
			\bold{z}^i &\sim \text{Bernoulli}(\bold{\pi^i})\\
			\bold{\pi^i} &= Sampler(h^i)
		\end{aligned}
	\end{equation}
	
	Another assumption is, following (IB) : 
	\begin{align}
		x\rightarrow h^1\rightarrow s^1 \rightarrow\dots \rightarrow h^H\rightarrow s^H \rightarrow \hat{y}
	\end{align} 
	is a markov chain.

	During the training, our goal is to maximize the mutual information of the pruned hidden states and the true label, i.e. $I(s^i;y)$, as well as control the mutual information before and after the pruning, i.e. $I(h^i;s^i)$. Added $\beta \ge 0$ for the tradeoff, we have the variational bottleneck loss function
	\begin{align}
		J_{IB} &= \sum_{i=1}^H [-I(s^i;y) + \beta I(s^i;h^i)] \label{infor}
	\end{align}

	However, the architecture of BERT does not facilitate tractable computation of (\ref{infor}). We adopt the variational approximation technique in \citep{VIB} to get its upper bound. 
	
	Let $q(y|s^i)$ be a variational approximation to $p(y|s^i)$ and $r(s^i) \sim N(0,1)$ to $p(s^i)$, now the upper bound of $-I(s^i;y) + \beta I(s^i;h^i)$ is 
	
	\begin{equation}
		\begin{aligned}
			&-E_{s^i \sim p(s^i | x),(x,y)\sim \mathcal{D}} [p(s^i|x)\log q(y|s^i)] + H(y)\\
			& \qquad + \beta E_{s^i \sim p(s^i | h^i)} [\log \dfrac{p(s^i|h_n^i)}{r(s^i)}] \label{layerinfor}
		\end{aligned}
	\end{equation}

	Please refer to appendix \ref{deduction} for the detailed derivation.

	\subsubsection{information bottleneck loss}
	Since here $p(s^i|x)$ represents the hidden states of x in the forward pass, and $q(y|s^i)$ equals the final classification output based on $s^i$, the first item in (\ref{layerinfor}) is equivalent to the cross entropy loss.

	Splitting the second item in (\ref{layerinfor}) into two parts:
	\begin{align}
		E_{s^i \sim p(s^i | h^i)} [\log p(s^i|h_n^i)] - E_{s^i \sim p(s^i | h^i)} [\log r(s^i)] \label{secondpart}
	\end{align}
	Given the training set $\{(x_n,y_n),n=1,\dots,N\}$, we estimate $p(s^i|x_n)=\delta_{s^i=s_n^i}$  where $s_n^i=Pruner(z_n^i,h_n^i),z_n^i = Sampler(h_n^i)$, and $h_n^i$ is the $i$th layer's hidden state of $x_n$ before entering the $Sampler_i$in the forward pass.Conditioned on $h_{n}^i$, $s^i$ and $z^i$ is one-to-one. 
	The former part of (\ref{secondpart}) therefore equals
	
	\begin{align}
		&\int \mathrm{d} s^i p(s^i | h_{n}^i) \log p(s^i | h_{n}^i)\nonumber \\
		&=-H(s^i | h_{n}^i)\\
		&=-H(z^i | h_{n}^i) \nonumber
	\end{align}
	where the masks 
	$
	z^i=(z_{1}^i, \ldots, z_{L}^i) \in\{0,1\}^{L}$ that are conditioned on $h_n^i$,
	are independent variables following
	\begin{align}
		z_{l}^i &\sim \text{Bernoulli}(\pi^i_{l}), l=1,2, \ldots, L
	\end{align}
	
	Therefore
	\begin{align}
		-H(z^i | h_{n}^i)&=\sum_{l=1}^{L}-H(z_{l}^i | h_{n}^i)\\
		&=\sum_{l=1}^{L} \pi^i_{l} \log \pi^i_{l} \nonumber \\
		& \qquad +(1-\pi^i_{l}) \log (1-\pi^i_{l}) \label{eq1}
	\end{align}
	
	But to get the second part in (\ref{secondpart}), which equals
	\begin{align}
		&\int \mathrm{d} s^i p(s^i | h_{n}^i) \log r(s^i)\nonumber \\
		=&E_{z\sim Bernoulli^{L}(\pi_l)}[r(s^i)] 
	\end{align}
	is computationally challenging, since the discrete probability space has $2^L$ outcomes.
	We simply estimate $p(s^i | h_{n}^i)$ with 
	\begin{equation}
		\begin{aligned}
			p(s^i | h_{n}^i) &\approx \delta_{s^i=s_n^i}\\
			s_n^i&=\text{diag} \{\pi_n^i\} h_n^i
		\end{aligned}
	\end{equation}
	in a forward propagation. 
	Hence now we get
	\begin{align}
		&\int \mathrm{d} s^i p(s^i | h_{n}^i) \log r(s^i) \nonumber
		\\=& \log\mathcal{N}(s_n^i;0,I) \\
		=& -\frac{L}{2} \log (2 \pi)-\frac{1}{2} \lVert s_n^i \rVert_F^2
		\nonumber
	\end{align}
	
	Finally, we can put everything together( and delete some constants) to get the following objective function, which we try to minimize:
	
	\begin{align}
		J_{IB}=\mathcal{L}_{ce}+\beta(\sum_{i=1}^H -H(z_i) + \frac{1}{2}\lVert s_n^i \rVert_F^2)
	\end{align}
	where $\mathcal{L}_{ce}$ is the cross entropy loss, and $H(z_i)$ is the entropy of the $i$th layer's token masks, computed by (\ref{eq1}) .
	
	In practice, we split the objective function into three losses, and scale them in terms of layers and size. The main training objective is 
	\begin{equation}
		\begin{aligned}
			\mathcal{L}=\mathcal{L}_{ce}+\gamma_{1}\mathcal{L}_{entropy}+\gamma_{2}\mathcal{L}_{norm}
		\end{aligned}
	\end{equation}
	
	\section{Experiments}
	\subsection{Setup}
	\paragraph{Datasets and metrics}
	To validate our approach, we apply it on four tasks of GLUE benchmark (\citealp{wang-etal-2018-glue}), including Stanford Sentiment Treebank (SST-2), Microsoft Research Paraphrase Corpus (MRPC), Question Natural Language Inference (QNLI) and Multi-Genre Natural Language Inference Matched (MNLI-m). The details are listed in table \ref{datasets}.
	\begin{table}[htp]
		\centering
		\begin{tabular}{llll}
			\hline
			\textbf{Dataset}& \textbf{\makecell[l]{Average \\ Length}} & \textbf{Task} & \textbf{metric}\\
			\hline
			MRPC&53& Paraphrase & F1 \\
			QNLI&51 & QA & acc. \\
			MNLI&39 & NLI & acc. \\
			SST2&32& Sentiment & acc. \\
			\hline
		\end{tabular}
		\caption{\label{datasets}
			Summary of evaluation datasets.
		}
	\end{table}
	\paragraph{Training Steps}
	We used the BERT$_\text{base}$ model as our base model and implemented a two-stage training process to create a compact and efficient model. In the first stage, we learn static pruning masks using a sparsity objective and a distillation loss. For more information about this stage, please refer to \citep{CoFi}. We kept training until arriving at a targeted pruning ratio $\in \{60\%,80\%,90\%,95\%\}$. Then we perform the token downsampling instead of the vanilla finetuning process. In specific, we first finetune the model with $\mathcal{L}_{ce}+\gamma_1 \mathcal{L}_{entropy}$ until convergence as a warmup. Then we add the $\mathcal{L}_{norm}$ to start the token sampling. The ratio of eliminated tokens is adjusted by $\gamma_1$ and $\gamma_2$. We set the seed to 42. (See Appendix \ref{hyper} for the hyperparameters setting)
	\paragraph{FLOPs and Parameters Calulation}
	We measure the inference FLOPs as a general measurement of the model's computational complexity, which allows us to assess models' speedups independent of their operating environment. We pad a batch of input sequences to the maximum length of the corresponding batch, with a batch size of 32. We calculate the FLOPs based on the model architecture as well as the inputs and the sampled masks. Then the FLOPs is averaged by tasks.
	When computing model parameters, following \citep{CoFi} and (Movement pruning: Adaptive sparsity by finetuning.), we exclude the parameters of the embedding layer.
	
	\paragraph{Baselines}
	We compare against several baselines, all of which are constructed based on BERT model:
	(1)  \textbf{TinyBERT}\citep{tinybert} and \textbf{DistillBERT}\citep{distilbert}: They are representative distillation models, both adopting general distillation and task-specific distillation. We also include TinyBERT without general distillation.
	(2) \textbf{CoFi} : The strong structured pruning model.
	(3) \textbf{PoWER-BERT}\citep{power} and \textbf{Transkimmer}\citep{Transkimmer}: Both of them are token pruning models. We did not include LTP(LTP) because it is constructed on RoBerta \citep{liu2020roberta}. 
	
	\begin{table*}[t]
		\centering
		\begin{tabular}{cccc|cccc}
			\hline
			Model        & params      &padding     & speedup                & MRPC(F1)                    & QNLI(acc)                   & SST2(acc) & MNLI(acc) \\ \hline
			BERT base& 100\% & -  & 1.0x  & 90.5  & 91.7 & 93.1 & 84.4      \\
			CoFi-s60 &40\%& - & 2.0x& 90.5 & 91.8& 93.0 & 85.3 \\
			CoFi-s95 &5\%& - & 8.2x& 85.6 & 86.1& 90.4 & 80.0 \\
			\hline
			PoWER-BERT & 100\%  & sequence    & 2.5x & 88.1  & 90.1 & 92.1 & 83.8      \\ 
			Transkimmer  & 100\%       & batch               & 2.3x                         & 89.1                        & 90.5                        & 91.1     & 83.2     \\ 
						CoFi-s80     & 20\% & - & 3.9x  & 88.6 & 90.1 & \textbf{92.5} & 83.9      \\
			\textbf{Infor-Coef-4x}  & 40\% & batch & 4.2x  & 90.5 & 90.6 &  91.2      &   84.5        \\
			\textbf{Infor-Coef-4x}  & 40\% & sequence & \textbf{5.0x}  & \textbf{90.5} & \textbf{90.6} &    91.2   &   \textbf{84.5}        \\ \hline
			TinyBERT$_4$  & 13\%      & -                  & 18.0x                        & 81.4                        & \textbf{86.7 }                       & 89.7      & 78.8      \\
			TinyBERT$_4$ w/o GD & 13\%   & - & 18.0x & 68.9 & 81.8 & 87.7 &  78.7  \\
			\textbf{Infor-Coef-16x} & 5\% &batch & 16.2x &  85.6 & 85.3 &90.1 &     79.1      \\ 
		\textbf{Infor-Coef-16x} & 5\%       &sequence                  & 18.0x &  \textbf{85.6} &     85.3 &\textbf{90.1} &    \textbf{79.1 }      \\ \hline
		\end{tabular}
		\caption{\label{baseline}
			Results on GLUE development set. GD denotes general distillation, which distills the student model on a large unlabeled data.
		}
	\end{table*}
	
	\begin{figure*}
		\centering
		\includegraphics[scale=0.39]{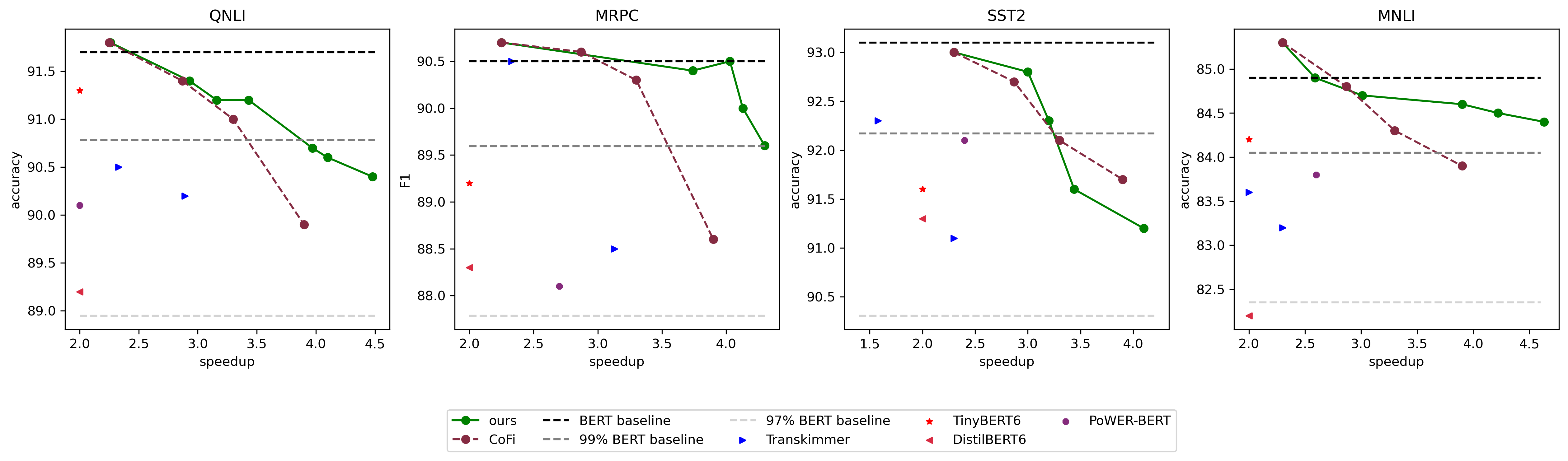}
		\caption{Accuracy-Speedup trade-off curve in a 2-4x speedup. We compress our model to the 60\% sparsity and apply token downsampling to different ratio. We then compare Infor-Coef(ours) against state-of-the-art pruning and distillation baselines.FLOPs speedup is analyzed using the padding strategy of "batch".}
		\label{tradobase}
	\end{figure*}

	\subsection{Overall Results}
	We begin by showing the overall results of our model in table \ref{baseline}.
	For a fair comparison, we train two models with a parameter size that equals CoFi-s60 or CoFi-90 so we could use the reported results in \citep{CoFi}. 
	
	Notably,  "padding" in  table \ref{baseline} stands for the padding strategy when implementing the token pruning models. According to \citep{LTP} when input sequences are padded to a fixed length, the results can be exaggerated because the pruning module tends to drop redundant padding tokens. However, we measured FLOPs using two types of padding strategies: "sequence," where sequences are padded to a fixed length according to PoWER-BERT \citep{power} (details are provided in Appendix \ref{padding}), and "batch," where sequences are padded according to the batch size.
	
	Our experiments demonstrate that our model achieves significant speedup with only a minor drop in accuracy (or F1 score on MRPC). We divide the model into three groups, with the backbone of BERT base and CoFi \citep{CoFi}. As demonstrated in the first group, on average our model achieves 5x speedup with less than 1\% accuracy degradation, and achieves 18x speedup with 5\% degradation. Compared to CoFi with the same weight compression rate, our models also experience less than 1\% accuracy drop but provide an acceleration in inference by 100\%. The substantial speedup does not depend on the model size, which is due to the orthogonality of the token downsampling strategy and the static pruning approach.  This allows us to achieve both a high level of compression and an acceleration in inference without sacrificing large model performance. 
		\begin{table*}[htp]
		\centering
		\begin{tabular}{c|ccccc} \hline
			& speedup & MRPC(F1)& QNLI(acc) & SST2(acc) &MNLI(acc)  \\ \hline
			Infor-Coef-4x& 4.2x& 90.6 & 90.6 &91.2 & 84.5\\
			-$\mathcal{L}_{entropy}$& 4.0x & 89.6(-1.0)&89.6(-0.9) &90.3(-0.9)&84.1(-0.4)\\
			-$\mathcal{L}_{norm}$ & 2.2x (-1.8x) &90.5&91.8 & 92.8 & 84.6\\ \hline
		\end{tabular}
		\caption{Ablation results on GLUE development set with 4.3x compression. We provide the results after removing the entropy loss and the norm loss.}
		\label{tab:lossabla}
	\end{table*}
	
	In the second group, we present the performance of our model with 40\% sparsity, namely Infor-Coef-4x. The comparative methods include the dynamic token pruning model baselines, which typically achieve 1.5-3 FLOPs speedup compared to the vanilla BERT model. Overall, we outperform the token pruning methods both in speedup ratio and accuracy. We also reimplement CoFi-s80, which denotes the CoFi model with 20\% weights and report the results, since it has a similar speedup ratio with Infor-Coef-4x. In the third group, we compare Infor-Coef-16x, which has a 16x-18x speedup, against TinyBERT$_4$ with or without general distillation. Infor-Coef-16x prunes 95\% of the model weights but has a competitive performance.  Empirically, our models outperform all the comparative models on three tasks in terms of speedup and accuracy.
	
	To showcase the flexibility and effectiveness of our models, we also compare their accuracy on GLUE development dataset to other methods while also measuring their inference speedup. These results are presented as tradeoff curves in Figure \ref{tradobase}. In particular, we outperformed CoFi on all tasks except SST2, which is consistent with the results presented in Table \ref{baseline}. Overall, our models achieve competitive performance when compared with other methods.
	
	We note that our model does not achieve the best performance on SST2 and QNLI in Table \ref{baseline} and Figure \ref{tradobase}. This is probably because the model is heavily influenced by similar training strategies and modules used in CoFi and Transkimmer. For instance, CoFi-s60 has a lower accuracy (86.1) than TinyBERT$_4$ (86.7) on QNLI. Although our model has higher compression rates compared to TinyBERT, it fails to surpass its performance when taking CoFi as our upper bound. Additionally, general distillation requires significant effort to pre-train a single model of a fixed size and computation, meaning that our strategy without pretraining could save substantial computation costs. Furthermore, SST2 has a shorter average length of 32 compared to other datasets in the GLUE benchmark (as shown in Table \ref{datasets}). According to \citealp{Transkimmer}, Transkimmer only achieves a 1.58x speedup on this dataset. This suggests that a small input size could handle the acceleration of token pruning methods.
	
	\subsection{Ablation Studies}
	\paragraph{Effects of Different Losses}
	To investigate the impact of different losses, we conduct experiments and present the results in Table \ref{tab:lossabla}. Although the improvement brought by entropy loss is not significant, we observed consistent improvements in the performance of our models across different GLUE datasets. The removal of the norm loss leads to the convergence toward a vanilla BERT model without dynamic accelerating. Theoretically, the entropy loss encourages the samplers to make a more "certain" decision, therefore it contributes to the stability of the performance. Including entropy loss only, however, may force the model to preserve all the tokens, leading to the vanilla model.
	
\begin{figure*}[h]
		\centering
		\includegraphics[scale=0.35]{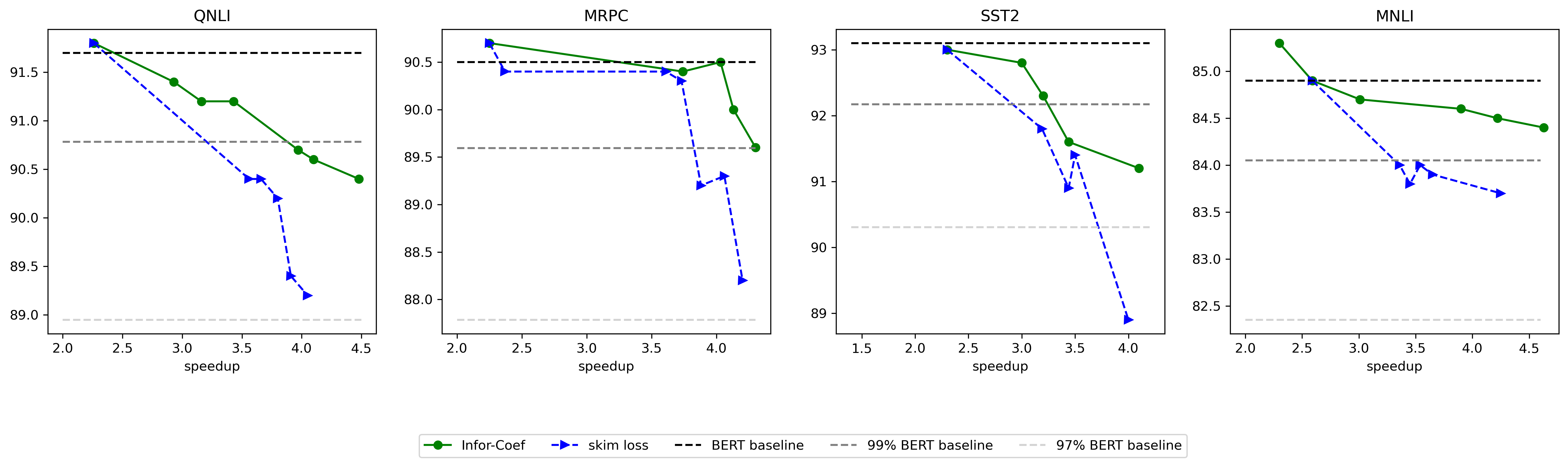}
		\caption{Accuracy-Speedup trade-off curve in a 2-4x speedup. We compress our model to the 60\% sparsity and apply token downsampling with the information bottleneck loss and skim loss. FLOPs speedup is analyzed using the padding strategy of "batch".}
		\label{skimloss}
	\end{figure*}

	As demonstrated in Figure \ref{skimloss}, we also compare our loss with the skim loss item in \citep{Transkimmer}, which is essentially the proportion of preserved tokens in each layer. We adopt the same hyperparameter setting in its original paper \citep{Transkimmer}. The FLOPs is calculated with the batch padding strategy, and all the models included are pruned with a 40\% parameter reduction. The trade-off curve suggests that our information bottleneck loss offers superior tradeoffs between accuracy and inference speedup when compared to skim loss.
	
	\paragraph{Acceleration Effects of static and dynamic pruning}

\begin{figure}[H]
	\centering
	\includegraphics[scale=0.35]{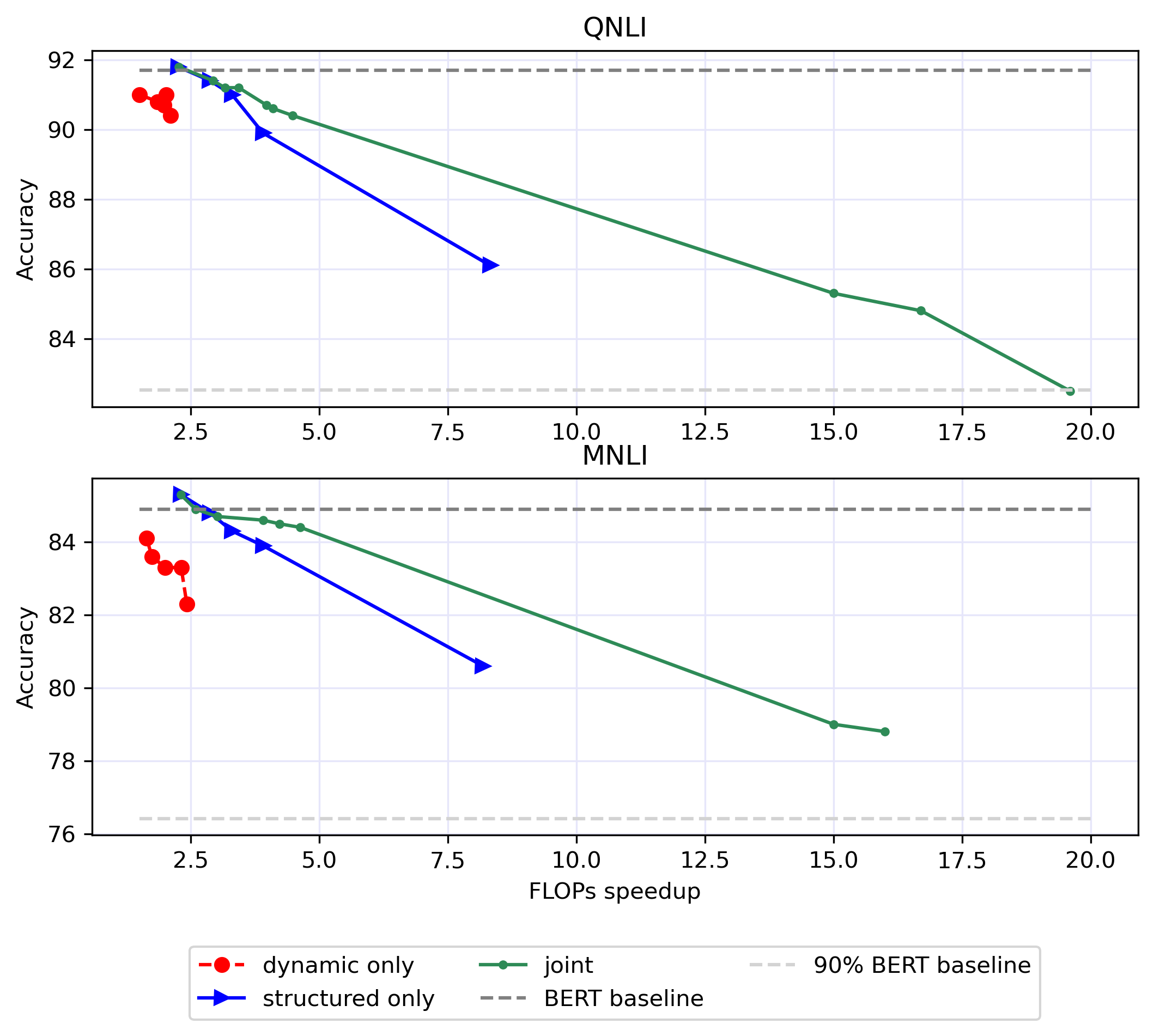}
	\caption{Trade-off results between accuracy and remaining FLOPs. We calculate the FLOPs ratio after pruning using batch padding. The "dynamic only", "structrured only" and "joint" mean conducting dynamic token downsampling, static pruning and both two strategy respectively.}
	\label{joint}
\end{figure}

	In this work, we propose a novel collaborative approach for model pruning that combines structural pruning and dynamic token pruning. We investigate the effects of this approach by systematically ablating different stages of the training process. Figure \ref{joint} provides a visual representation of our proposed approach.

	The figure demonstrates that the joint pruning outperforms the dynamic token downsampling significantly,  having both superior FLOPs compression and accuracy retaining. The dynamic downsampling only gets 1.5-2.5x FLOPs reduction without a large accuracy sacrifice, while our proposed method could reduce the FLOPs by 80\%.  Furthermore, the performance of joint pruning not only exceeds structured pruning but also provides a larger range of speedup. Compared with structured pruning which prunes 95\% parameters to get an approximately 10x speedup, Infor-Coef reaches a speedup ratio of larger than 17x, showing significant flexibility. 
\section{Conclusion}

In this paper, we propose a model acceleration approach for large language models that incorporates dynamic pruning and static pruning, optimized by the information bottleneck loss. Our models selectively and adaptively allocate computation on different inputs and hidden states, resulting in a slenderized and efficient subnetwork. We also introduced a novel information bottleneck-based training strategy that outperforms the vanilla $l0$ norm-like loss for dynamic token reduction. Our empirical results demonstrate that our approach can achieve over 16x speedup while maintaining 95\% performance. We conclude that different pruning methods are well-adaptable to each other through task-specific fine-tuning, and we hope that our work will inspire future research in the context of pruning large language models.

	\newpage

	\bibliography{mainbody}
	\bibliographystyle{acl_natbib}

	\newpage
	\newpage
	\appendix
	
	\section{Derivation of Informtion Bottleneck Upper Bound}
	\label{deduction}
	Using that Kullback Leibler divergence is always positive, we have		
	\begin{align}
		I(s^i ; y) &=\int \mathrm{d} s^i \mathrm{d} y p(s^i, y) \log \frac{p(y | s^i)}{p(y)} \nonumber \\
		&\geq \int \mathrm{d} s^i \mathrm{d} y p(s^i, y) \log \frac{q(y | s^i)}{p(y)}
	\end{align}
	
	Given the training set $\{(x_n,y_n),n=1,\dots,N\}$, we estimate $p(s^i|x_n)=\delta_{s^i=s_n^i}$  where $s_n^i=Pruner(z_n^i,h_n^i),z_n^i = Sampler(h_n^i),h_n^i$ is the $i$th layer's hidden state of $x_n$ before entering the $Sampler_i$in the forward pass.
	Leveraging our Markov assumption,
	\begin{align}
		p(y, s^i) &= \int \mathrm{d}x p(x, y, s^i)\nonumber \\
		& = \int \mathrm{d} x p(x)p(s^i|x)p(z|x)
	\end{align}
	We can rewrite the mutual information lower bound
	\begin{align}
		I(s^i;y)& \ge \int \mathrm{d} x d s^i \mathrm{d} y p(s^i | x) p(y | x) \log q(y | s^i) \nonumber \\
		& \quad-H(y)   \\ 
		& \approx \frac{1}{N} \sum_{n=1}^{N} \int \mathrm{d} s^i p(s^i | x_{n}) \log q(y_{n} | s^i) \nonumber \\
		& \quad -\text { constant }
	\end{align}
	
	Since here $q(y|s^i)$ equals the final classification output based on $s^i$, it is equivalant to minimize the cross entropy loss.
	
	For the second mutual information item, we let $r(s^i) \sim N(0,1)$ be a variational approximation to $p(s^i)$. Using  Kullback Leibler divergence again, we have	
	
	\begin{align}
		I(s^i ; h^i) & =\int \mathrm{d} s^i \mathrm{d} h^i p(s^i, h^i) \log \frac{p(s^i | h^i)}{p(s^i)} \nonumber \\
		& =\int \mathrm{d} s^i \mathrm{d} h^i p(h^i) p(s^i | h^i) \log \frac{p(s^i | h^i)}{p(s^i)}  \\
		& \leq \int \mathrm{d} s^i \mathrm{d} h^i p(h^i) p(s^i | h^i) \log \frac{p(s^i | h^i)}{r(s^i)} \nonumber
	\end{align}
	Given the training dataset $\{(x_i,y_i)\}_{i=1}^N$, the upper bound can be approximated as
	\begin{align}
		& \int \mathrm{d} s^i \mathrm{d} h^i p(h^i) p(s^i | h^i) \log \frac{p(s^i | h^i)}{r(s^i)} \nonumber \\
		& \approx \frac{1}{N} \sum_{n=1}^{N} \int \mathrm{d} s^i p(s^i | h_{n}^i) \log \frac{p(s^i | h_{n}^i)}{r(s^i)}  \\
		& =\frac{1}{N} \sum_{n=1}^{N} \int [\mathrm{d} s^i p(s^i | h_{n}^i) \log p(s^i | h_{n}^i)\nonumber \\
		&\qquad-\mathrm{d} s^i p(s^i | h_{n}^i) \log r(s^i)] \nonumber
	\end{align}
	
	\section{Fixed Padded length}
	\label{padding}
	
	Following \citep{power}, we pad the inputs into a fixed length depending on different datasets.

	\begin{table}[H]
		\centering
		\begin{tabular}{|c|c|}
			\hline
			Dataset & Length \\ \hline
			MRPC & 128 \\
			MNLI & 128 \\
			QNLI & 128 \\
			SST2 & 64 \\ \hline
		\end{tabular}
	\caption{Padding length on the evaluation dataset.}
	\end{table}
	
	\section{Training Parameters}
	\label{hyper}
	
We provide the hyperparameters used in our experiments as a reference for reimplementing our method.  However, we acknowledge that the results may differ slightly depending on various factors such as the hardware devices and package versions.
\begin{table}[H]
\begin{tabular}{c|cc}
	\hline
	& MRPC & QNLI \\ \hline
	batch size & 32 & 32 \\
	learning rate & 1e-5,2e-5,5e-5 &1e-5,2e-5\\
	norm coef & 5e-4,6e-4 & 5e-4,7e-4 \\
	entropy coef & 0,5e-4 & 3e-4,4e-4 \\
	epoch & 10 & 5 \\ \hline
	 & MNLI & SST2 \\ \hline
	 batch size  & 32 & 32 \\
	 learning rate  & 1e-5,2e-5 & 1e-5,2e-5 \\
	 norm coef & 5e-4,4e-4 & 5e-4,4e-4 \\
	 entropy coef &  4e-4 & 5e-4,6e-4 \\
	 epoch & 3 & 10 \\ \hline
\end{tabular}
	 \caption{Hyper parameters setting.}
\end{table}

\end{document}